\documentclass[conference]{IEEEtran}
\IEEEoverridecommandlockouts
\usepackage{cite}
\usepackage{pbalance} %balances the last page automatically
\usepackage{tabularx}
\usepackage{amsmath,amssymb,amsfonts}
\usepackage[table,xcdraw]{xcolor}
\usepackage{algorithmic}
\usepackage{graphicx}
\usepackage{textcomp}
\usepackage{cleveref} % for \cref command
\usepackage{url}
\usepackage{adjustbox}
\usepackage{amsmath}
\usepackage{algorithm}
\usepackage{algorithmic}
\usepackage{float} % for the [H] option
\usepackage{xcolor}
\def\BibTeX{{\rm B\kern-.05em{\sc i\kern-.025em b}\kern-.08em
    T\kern-.1667em\lower.7ex\hbox{E}\kern-.125emX}}
\begin{document}

% \title{Real-Time Distracted Driving Detection Using YOLOv8: An Innovative Approach for High-Accuracy and Efficient Classification}

\title{P-YOLOv8: Efficient and Accurate Real-Time Detection of Distracted Driving

\thanks{The authors would like to thank the following funding agencies: NSF grant 2219680.}

}

\author{
    \IEEEauthorblockN{Mohamed R. Elshamy\IEEEauthorrefmark{1}, 
    Heba M. Emara\IEEEauthorrefmark{2}, 
    Mohamed R. Shoaib\IEEEauthorrefmark{3}, 
    Abdel-Hameed A. Badawy\IEEEauthorrefmark{1}}
    
    \IEEEauthorblockA{\IEEEauthorrefmark{1}Klipsch School of ECE, New Mexico State University, Las Cruces, NM 88003, United States \\
    %Email: \{elshamy, badawy\}@nmsu.edu
    }
    \IEEEauthorblockA{\IEEEauthorrefmark{2}Dept.\ of Electronics and Communications Engineering, Pyramids Higher Inst.\ for Engineering \& Technology, Egypt\\
    %Email: heba.emara@el-eng.menofia.edu.eg
    }
    \IEEEauthorblockA{\IEEEauthorrefmark{3}College of Computing and Data Science, Nanyang Technological University, Singapore \\
    %Email: Mohamedr003@e.ntu.edu.sg
    }
    \IEEEauthorblockA{\IEEEauthorrefmark{1}\{elshamy, badawy\}@nmsu.edu, \IEEEauthorrefmark{2}heba.emara@el-eng.menofia.edu.eg, \IEEEauthorrefmark{3} Mohamedr003@e.ntu.edu.sg}
}
 
\maketitle

\begin{abstract}

Distracted driving is a critical safety issue that leads to numerous fatalities and injuries worldwide. This study addresses the urgent need for efficient and real-time machine learning models to detect distracted driving behaviors. Leveraging the Pretrained-YOLOv8 (P-YOLOv8) model, a real-time object detection system is introduced, optimized for both speed and accuracy. This approach addresses the computational constraints and latency limitations commonly associated with conventional detection models. The study demonstrates P-YOLOv8's versatility in both object detection and image classification tasks using the \textit{Distracted Driver Detection} dataset from state farm, which includes $22,424$ images across ten behavior categories. Our research explores the application of P-YOLOv8 for image classification, evaluating its performance compared to deep learning models such as VGG16, VGG19, and ResNet. Some traditional models often struggle with low accuracy, while others achieve high accuracy but come with high computational costs and slow detection speeds, making them unsuitable for real-time applications. P-YOLOv8 addresses these issues by achieving competitive accuracy with significant computational cost and efficiency advantages. In particular, P-YOLOv8 generates a lightweight model with a size of only 2.84 MB and a lower number of parameters, totaling $1,451,098$, due to its innovative architecture. It achieves a high accuracy of $99.46\%$ with this small model size, opening new directions for deployment on inexpensive and small embedded devices using Tiny Machine Learning (TinyML). The experimental results show robust performance, making P-YOLOv8 a cost-effective solution for real-time deployment. This study provides a detailed analysis of P-YOLOv8's architecture, training, and performance benchmarks, highlighting its potential for real-time use in detecting distracted driving.

\end{abstract}


\section{Introduction}
Distracted driving poses significant risks to road safety, resulting in approximately $3,142$ deaths and $424,000$ injuries in the United States in $2019$, according to the Centers for Disease Control and Prevention (CDC)~\cite{cdc2019}. This averages nine deaths per day, and $20\%$ of the deceased were pedestrians or cyclists, highlighting the widespread impact beyond vehicle occupants. The urgent need to mitigate these risks emphasizes the importance of effective methods to detect distracted driving behaviors. 

Machine learning (ML) can facilitate the automatic identification of driver inattention, crucial to improving road safety and transforming how insurance companies assess driving behaviors. By monitoring habits through dashboard-mounted cameras, insurers can adjust premiums based on driver attentiveness, offering lower rates to safer drivers. Although numerous studies have introduced efficient algorithms for this problem, most are computationally expensive and produce large model sizes, focusing mainly on accuracy without adequate consideration of detection speed~\cite{Masood2020,10.1007/978-981-19-6634-7_47, ResNet,9933740,9728846,9781938}. In contrast, the proposed P-YOLOv8 (You Only Look Once, version 8) algorithm optimizes accuracy and speed, resulting in a more efficient and scalable solution for real-time applications~\cite{ultralytics2023classify}.

P-YOLOv8 achieves superior performance through several key innovations. Its streamlined architecture reduces parameters and computational overhead while maintaining accuracy, contrasting starkly with traditional deep learning models such as VGG16, VGG19, and ResNet~\cite{10334988, Shoaib_2021}. The effective use of anchor boxes by P-YOLOv8 and improved bounding box prediction strategies further boost detection precision and speed. Furthermore, real-time performance is achieved through batch normalization and efficient memory usage during inference~\cite{10334988}. The YOLO model, originally designed for object detection, has been adapted for image classification. P-YOLOv8 represents a significant evolution in the YOLO series, incorporating enhancements that improve performance, flexibility, and efficiency, making it suitable for applications requiring real-time processing~\cite{ultralytics2023classify,10385083,10425093,10550783}.

This study focuses on the P-YOLOv8 models, specifically the variant yolov8n-cls.pt, optimized for efficient image classification tasks. These models assign a single class label to an entire image, accompanied by a confidence score, which is advantageous for applications where determining the overall class is sufficient without identifying specific objects. Using the State Farm "Distracted Driver Detection" dataset~\cite{statefarm2017}, which consists of $22,424$ images in ten behavior categories, the aim is to improve detection speed and classification of potentially dangerous activities. The dataset includes various forms of driver distractions, such as texting and talking on the phone. Experimental results demonstrate that P-YOLOv8 achieves competitive accuracy in image classification tasks while offering significant advantages in speed and computational efficiency, making it a viable alternative to traditional deep learning classification models.

The rest of the paper is organized as follows: Section II reviews related work. Section III details the dataset and the proposed algorithm. Section IV presents experimental results and a discussion. Section V concludes the paper.

\section{Related Work}
Distracted driving detection is crucial for road safety, with various models developed to identify driver distractions, each showing unique strengths and limitations. 
Hossain~\textit{et al.}~\cite{HOSSAIN2022200075} proposed an automatic driver distraction detection method using deep convolutional neural networks, achieving a maximum accuracy of $99.98\%$ with the MobileNet v2 model on the State Farm dataset. Despite its high accuracy, the relatively large parameter count of the model (3.5 million) may hinder deployment on resource-constrained devices. In comparison, the VGG-16 model has $138.3$ million parameters, while the ResNet50 model has $25,6$ million parameters.

Sajid \textit{et al.}~\cite{9662336} developed an efficient deep learning framework for distracted driver detection using the State Farm dataset, reporting that the EfficientDet-D3 model achieved a maximum mean average precision (MAP) of $99.16\%$. However, its complexity poses challenges for deployment on constrained devices, as the model requires extensive training epochs to achieve high accuracy. Bahari~\textit{et al.}~\cite{ResNet} also reported high accuracy with the ResNet50 model ($94\%$), but its large size and complexity can limit practical deployment. 
Aljasim \textit{et al.}~\cite{s22051858} created an ensemble model combining ResNet50 and VGG16, achieving a maximum accuracy of $92\%$. However, this model's complexity and parameter count (ResNet50: 23 million and VGG16: 138 million) could affect deployment on resource-constrained devices. 

Masood \textit{et al.}~\cite{Masood2020} reported that the VGG16 model achieved a maximum accuracy of $99.57\%$, but its high parameter count (~138 million) presents challenges for resource-constrained environments. 
Fang \textit{et al.}~\cite{9965124} achieved a maximum accuracy of $99.57\%$ using the Vision Transformer (ViT) model with transfer learning, but its complexity poses similar deployment challenges. Subbulakshmi \textit{et al.}~\cite{10353386} presented an ensemble model with a maximum accuracy of $97.5\%$, but its large parameter sizes (\textit{e.g.}, NasNet-A Large: ~88 million, ResNeXt-101: ~44 million) complicate deployment.

Li~\textit{et al.}~\cite{9913644} developed a framework achieving maximum accuracies of $99. 92\%$ (AlexNet), $100\%$ (VGG16) and $99. 99\%$ (ResNet18), but all exhibited complexity and substantial parameter counts (\textit{e.g.}, AlexNet: 217 million, VGG16: 491 million). Abbas \textit{et al.}~\cite{9318087} reported that the optNet-50 model achieved a maximum accuracy of $98\%$, but its parameter count (317.4 million) complicates the deployment. Li~\textit{et al.}~\cite{10295832} presented a hybrid convolutional transformer model (MViTCNet) achieving an accuracy of $91.04\%$ with a more manageable parameter count of $1.36$ million, offering a favorable balance between accuracy and computational efficiency.

Detecting distracted driving behaviors is critical for road safety, as various models demonstrate high accuracy but often face deployment challenges on resource-constrained devices due to complexity and large parameter sizes. For example, MobileNet v2 achieves $99.98\%$ precision but has 3.5 million parameters~\cite{HOSSAIN2022200075}. VGG16~\cite{Masood2020} and ResNet50~\cite{ResNet} have 138 million and $25.6$ million parameters, respectively. Models such as EfficientDet-D3~\cite{9662336}, ResNet50~\cite{ResNet}, and ensemble models~\cite{s22051858} also report high accuracy but require significant computational resources. To address this, we propose leveraging P-YOLOv8, which balances accuracy and computational efficiency, making it suitable for real-time applications on devices with limited resources. Our approach aims to enhance real-world applicability by offering a superior trade-off.

\section{Material and Method}
\subsection{Dataset Description} 

\begin{figure}[htb]
\vspace{-0.3cm} % Adjust this value as needed to remove space
\centering
\includegraphics[width=.9\linewidth]{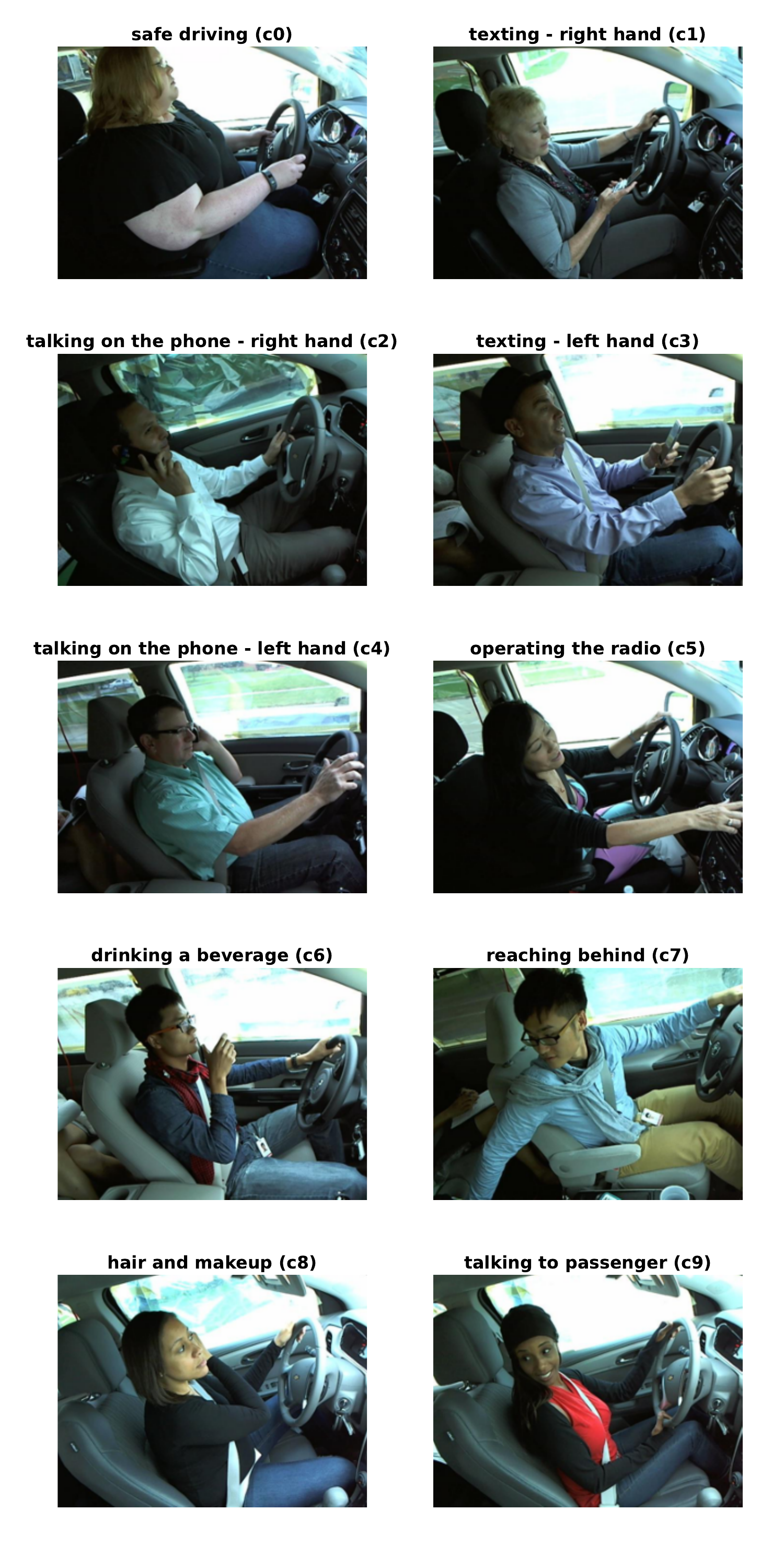}
\vspace{-0.6cm} % Adjust this value as needed to
\caption{Examples of different driving behaviors captured in the dataset.}
\vspace{-0.6cm} % Adjust this value as needed to
\label{fig:class_images}
\end{figure}

To evaluate the proposed method (P-YOLOv8), we utilized the State Farm dataset~\cite{statefarm2017}, which contains $22,424$ images classified into ten distinct classes. Each image in the dataset is presented as a \(640 \times 480\) RGB image. The first category represents safe driving, while the remaining nine categories relate to various forms of distracted driving (\textit{e.g.}, texting, talking on the phone) as illustrated in Figure~\ref{fig:class_images}. The distribution of the images across these classes is shown in Table~\ref{tab:class_numbers}. The dataset was divided into training, validation, and test sets, with $70\%$ of the images allocated for training, $15\%$ for validation, and 15\% for testing.

\begin{table}[htb]
\centering
\vspace{-0.5cm}
\caption{Class names and their corresponding numbers}
\label{tab:class_numbers}
\begin{tabularx}{\columnwidth}{|c|X|c|}
\hline
Class & Name & Number of Images \\
\hline
c0 & Safe driving & 2489 \\
c1 & Texting - right hand & 2267 \\
c2 & Talking on the phone - right hand & 2317 \\
c3 & Texting - left hand & 2346 \\
c4 & Talking on the phone - left hand & 2326 \\
c5 & Operating the radio & 2312 \\
c6 & Drinking a beverage & 2325 \\
c7 & Reaching behind & 2002 \\
c8 & Hair and makeup & 1911 \\
c9 & Talking to passenger & 2129 \\
\hline
\end{tabularx}
\end{table}

\subsection{The Proposed Algorithm}
P-YOLOv8, the latest iteration of the YOLO series by Ultralytics, represents a significant advance in computer vision. Building on the success of its predecessors, YOLOv8 introduces enhancements that improve performance, flexibility, and efficiency. It supports a wide range of vision AI tasks, including object detection, segmentation, pose estimation, tracking, and classification, making it a versatile tool for various applications~\cite{10385083,10425093,10550783,10453737}. Table~\ref{tab:yolov8} summarizes the YOLOv8 pre-trained classification models. Detection, segmentation, and pose estimation models are trained on the COCO data set, while classification models use the ImageNet dataset~\cite{ultralytics2024imagenet}. These models, trained in a large number of labeled images, exhibit strong generalization capabilities for image recognition tasks. This study employs the P-YOLOv8 algorithm for classification on a NVIDIA RTX-A4000 GPU.

%P-YOLOv8, the latest iteration of the YOLO series developed by Ultralytics, represents a significant advancement in the field of computer vision. Building upon the success of its predecessors, YOLOv8 incorporates numerous enhancements that improve its performance, flexibility, and efficiency. The model is designed to support a comprehensive range of vision AI tasks, including object detection, segmentation, pose estimation, tracking, and classification. This versatility makes YOLOv8 a powerful tool for various applications and domains, offering state-of-the-art solutions for contemporary vision-based challenges~\cite{10385083,10425093,10550783,10453737}.
%Table~\ref{tab:yolov8} presents YOLOv8's pre-trained classification models. While the detection, segmentation, and pose estimation models have been trained on the COCO dataset, the classification models have been trained on the ImageNet dataset~\cite{ultralytics2024imagenet}. Furthermore, the table provides information about various versions of the YOLOv8 model that have been pre-trained for image classification tasks. These models are specifically trained on the ImageNet dataset~\cite{ultralytics2024imagenet}, which contains a vast number of labeled images across many categories, allowing the models to learn and generalize well for image recognition. This paper uses the P-YOLOv8 algorithm for classification with NVIDIA RTX-A4000 GPU.

\begin{table*}[htb]
\centering
\caption{YOLOv8 Classification Models Performance}
\label{tab:yolov8}
\begin{tabular}{|l|c|c|c|c|c|c|c|}
\hline
Model & size (pixels) & acc top1 & acc top5 & Speed CPU ONNX (ms) & Speed A100 TensorRT (ms) & params (M) & FLOPs (B) at 640 \\
\hline
YOLOv8n-cls & 224 & 69.0 & 88.3 & 12.9 & 0.31 & 2.7 & 4.3 \\
YOLOv8s-cls & 224 & 73.8 & 91.7 & 23.4 & 0.35 & 6.4 & 13.5 \\
YOLOv8m-cls & 224 & 76.8 & 93.5 & 85.4 & 0.62 & 17.0 & 42.7 \\
YOLOv8l-cls & 224 & 76.8 & 93.5 & 163.0 & 0.87 & 37.5 & 99.7 \\
YOLOv8x-cls & 224 & 79.0 & 94.6 & 232.0 & 1.01 & 57.4 & 154.8 \\
\hline
\end{tabular}
\end{table*}

Table~\ref{tab:yolov8} summarizes the P-YOLOv8 models: YOLOv8n-cls, YOLOv8s-cls, YOLOv8m-cls, YOLOv8l-cls, and YOLOv8x-cls, which vary in size and complexity, affecting performance and speed. All models use a uniform input size of 224 pixels. The models' top-1 and top-5 accuracies on the ImageNet validation set improve with complexity, ranging from 69.0\% and 88.3\% for YOLOv8n-cls to 79.0\% and 94.6\% for YOLOv8x-cls. Inference times vary, with YOLOv8n-cls processing an image in 12.9 ms on a CPU and 0.31 ms on an NVIDIA A100 GPU, while YOLOv8x-cls requires up to 232 ms on a CPU. The complexity of the model, indicated by the number of parameters and FLOPs, ranges from 2.7 million parameters and 4.3 billion FLOPs for YOLOv8n-cls to 57.4 million parameters and 154.8 billion FLOPs for YOLOv8x-cls. This study focuses on the variant YOLOv8n-cls.pt, optimized for efficient image classification, assigning a single class label with a confidence score, ideal for tasks that require general class determination~\cite{7780460,8100173}.

%The Table~\ref{tab:yolov8} presents different versions of the P-YOLOv8 models: YOLOv8n-cls, YOLOv8s-cls, YOLOv8m-cls, YOLOv8l-cls, and YOLOv8x-cls, varying in size and complexity, which impact their performance and speed. All models use a uniform input size of 224 pixels, facilitating consistent feature extraction. The models' top-1 and top-5 accuracies on the ImageNet validation set improve with complexity, with YOLOv8n-cls achieving 69.0\% and 88.3\%, and YOLOv8x-cls achieving 79.0\% and 94.6\%, respectively. Inference speeds vary, with simpler models like YOLOv8n-cls processing an image in 12.9 ms on a CPU and 0.31 ms on an NVIDIA A100 GPU, whereas the more complex YOLOv8x-cls takes up to 232 ms on a CPU. The number of parameters and FLOPs, indicating model complexity, range from 2.7 million and 4.3 billion for YOLOv8n-cls to 57.4 million and 154.8 billion for YOLOv8x-cls, respectively. This study focuses on the YOLOv8 models, specifically the YOLOv8n-cls.pt variant, optimized for efficient image classification tasks, assigning a single class label to an entire image with a confidence score, ideal for applications needing overall class determination without identifying object locations or shapes within the image~\cite{7780460,8100173}.

P-YOLOv8 is a state-of-the-art real-time object detection system that improves its predecessors with a powerful backbone (CSPDarknet53)~\cite{10118431}, a robust neck (PANet), and an efficient prediction head. The backbone extracts features via convolutional layers with batch normalization and activation functions, while the neck aggregates features for the head to predict bounding boxes, objectness scores, and class probabilities, enabling high-speed single-pass processing.

The output of a convolutional layer is given by:
\begin{equation}
    F_{l+1} = \sigma ( W_l * F_l + b_l )
\end{equation}
where \( F_{l} \) is the feature map, \( W_{l} \) are the weights, \( * \) denotes convolution, \( b_{l} \) is the bias, and \( \sigma \) is the activation function.

Bounding box predictions are represented as:
\begin{equation}
    \mathbf{b} = [x, y, w, h]
\end{equation}
where \( x \) and \( y \) are the center coordinates, and \( w \) and \( h \) are the dimensions. Class probabilities are calculated using:
\begin{equation}
    \hat{y}_j = \frac{\exp(z_j)}{\sum_{k=1}^{K} \exp(z_k)}
\end{equation}
where \( \hat{y}_j \) is the predicted probability for class \( j \).
P-YOLOv8 can also be adapted for classification by utilizing its backbone and head to produce class probabilities through feature extraction, grouping, flattening, and fully connected layers. Its efficiency and real-time performance stem from a unified architecture and optimized inference, while accuracy is enhanced by advanced feature extraction and data augmentation techniques~\cite{10353386}. The detailed algorithm structure is provided in~\Cref{alg:pyolov8_classification}. It describes a method for detecting distracted driving using a P-YOLOv8 model $M$. The algorithm includes data preprocessing (loading, resizing, normalizing, augmenting, and splitting the dataset), model training (loading the pre-trained model, fine-tuning, training, validating, and updating model parameters with backpropagation), and model evaluations. This approach leverages the robust YOLOv8 architecture to ensure the trade-off between accuracy and computational cost.

\begin{algorithm}
\caption{P-YOLOv8-Based Distracted Driving Detection}
\label{alg:pyolov8_classification}
\begin{algorithmic}[1]
\REQUIRE Dataset $D$ with images of driving behaviors, pre-trained YOLOv8 model $M$ (yolov8n-cls.pt)
\ENSURE Dataset Structure for YOLO Classification Tasks~\cite{ultralytics2023classify}

\STATE \textbf{Data Preprocessing:}
\STATE Load the dataset $D$
\STATE Resize images to 224 × 224 pixels
\STATE Normalize pixel values
\STATE Perform data augmentation (rotation, flipping, scaling)
\STATE Split the dataset into training, validation, and test sets

\STATE \textbf{Model Training:}
\STATE Load the pretrained YOLOv8 model $M$ (\textit{e.g.}, yolov8n-cls.pt)
\STATE Fine-tune $M$ on the training set
\FOR{each epoch}
    \STATE Train the model $M$ on the training set
    \STATE Validate the model $M$ on the validation set
    \STATE Compute training and validation loss:
    \begin{equation}
        L = \sum_{i=1}^{N} y_i \log(\hat{y}_i) + (1 - y_i) \log(1 - \hat{y}_i)
    \end{equation}
    \STATE Update model parameters using backpropagation:
    \begin{equation}
        \theta \leftarrow \theta - \eta \nabla_\theta L
    \end{equation}
\ENDFOR

\STATE \textbf{Model Evaluations}
\end{algorithmic}
\end{algorithm}

\section{Results and Discussion}
The results of the P-YOLOv8 models in the provided dataset elucidate the trade-offs between accuracy and computational cost. The key metrics summarizing the performance include an impressive overall accuracy of \(99.464\%\) and detailed precision, recall, and F1 score for each class, as shown in Table~\ref{table:metrics}. Furthermore, the size of the model is remarkably tiny at $2.84 MB$, featuring a lower number of parameters, totaling $1,451,098$, due to its innovative architecture.
The classification metrics for the P-YOLOv8 model, as presented in Table~\ref{table:metrics}, demonstrate its high performance in multiple classes. The model achieves nearly perfect precision, recall, and F1 scores for most classes, with precision and recall values consistently close to or at 1.000000 for classes c1, c2, c5, c6, and c7. The average precision, recall, and F1 score are all above 0.99, specifically 0.994239, 0.994648, and 0.994422, respectively. These results highlight the model's exceptional ability to accurately classify different behaviors, maintaining balanced performance and robust reliability. Additionally, the performance of class c8, while slightly lower than others, still remains high, indicating the overall effectiveness and precision of the model in classification tasks.

\begin{table}[tb]
\vspace{-0.4cm}
\caption{Classification Metrics}
\vspace{-0.1cm}
\centering
\begin{tabularx}{\columnwidth}{|c|X|X|X|}
\hline
\textbf{Class} & \textbf{Precision} & \textbf{Recall} & \textbf{F1 Score} \\
\hline
c0 & 0.991957 & 0.991957 & 0.991957 \\
c1 & 1.000000 & 1.000000 & 1.000000 \\
c2 & 1.000000 & 1.000000 & 1.000000 \\
c3 & 0.997151 & 0.997151 & 0.997151 \\
c4 & 0.997126 & 0.994269 & 0.995696 \\
c5 & 1.000000 & 1.000000 & 1.000000 \\
c6 & 1.000000 & 0.991453 & 0.995708 \\
c7 & 0.996667 & 1.000000 & 0.998331 \\
c8 & 0.972028 & 0.996416 & 0.984071 \\
c9 & 0.987461 & 0.975232 & 0.981308 \\
\hline
Average & 0.994239 & 0.994648 & 0.994422 \\
\hline
\end{tabularx}
\label{table:metrics}
\end{table}

\begin{figure}%[htbp]
\vspace{-0.2cm} % Adjust this value as needed to remove space
\centering
\includegraphics[width=0.9\linewidth]{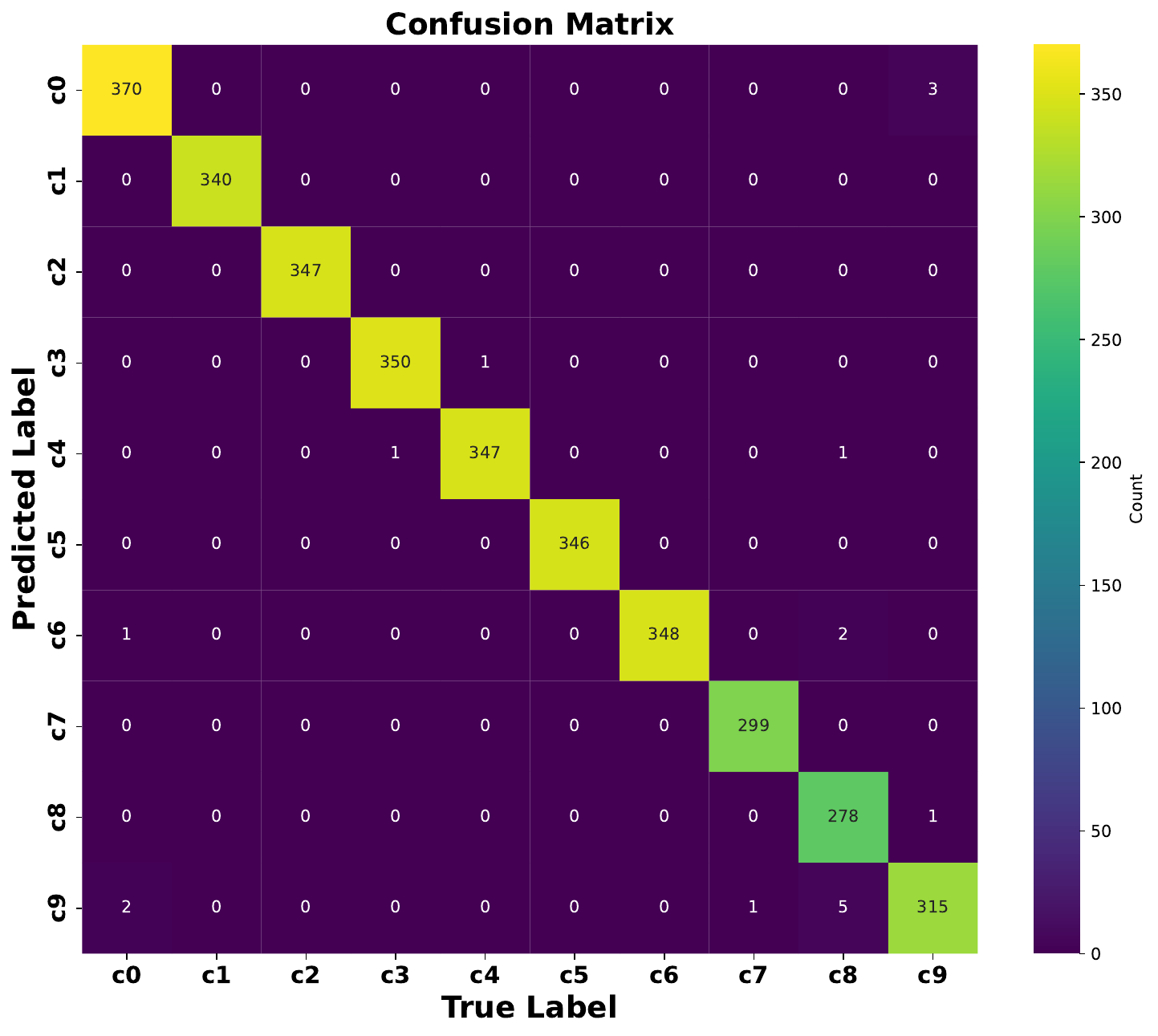}
\vspace{-0.2cm}
\caption{Confusion Matrix Visualization. The matrix shows the number of correct and incorrect predictions for each class.}
\label{fig:confusion matrix}
\end{figure}

The confusion matrix was also used to gain a more granular view of our model performance. The confusion matrix provides a detailed breakdown, indicating the number of correct and incorrect predictions for each class. As shown in the confusion matrix visualization in Figure~\ref{fig:confusion matrix}, the high diagonal values indicate that the model accurately predicts the majority of instances in all classes.

\begin{figure}%[htbp]
\centering
\vspace{-0.5cm} % Adjust this value as needed to 
\includegraphics[trim={7cm 0 2.2cm 0},clip,width=1.1\linewidth]{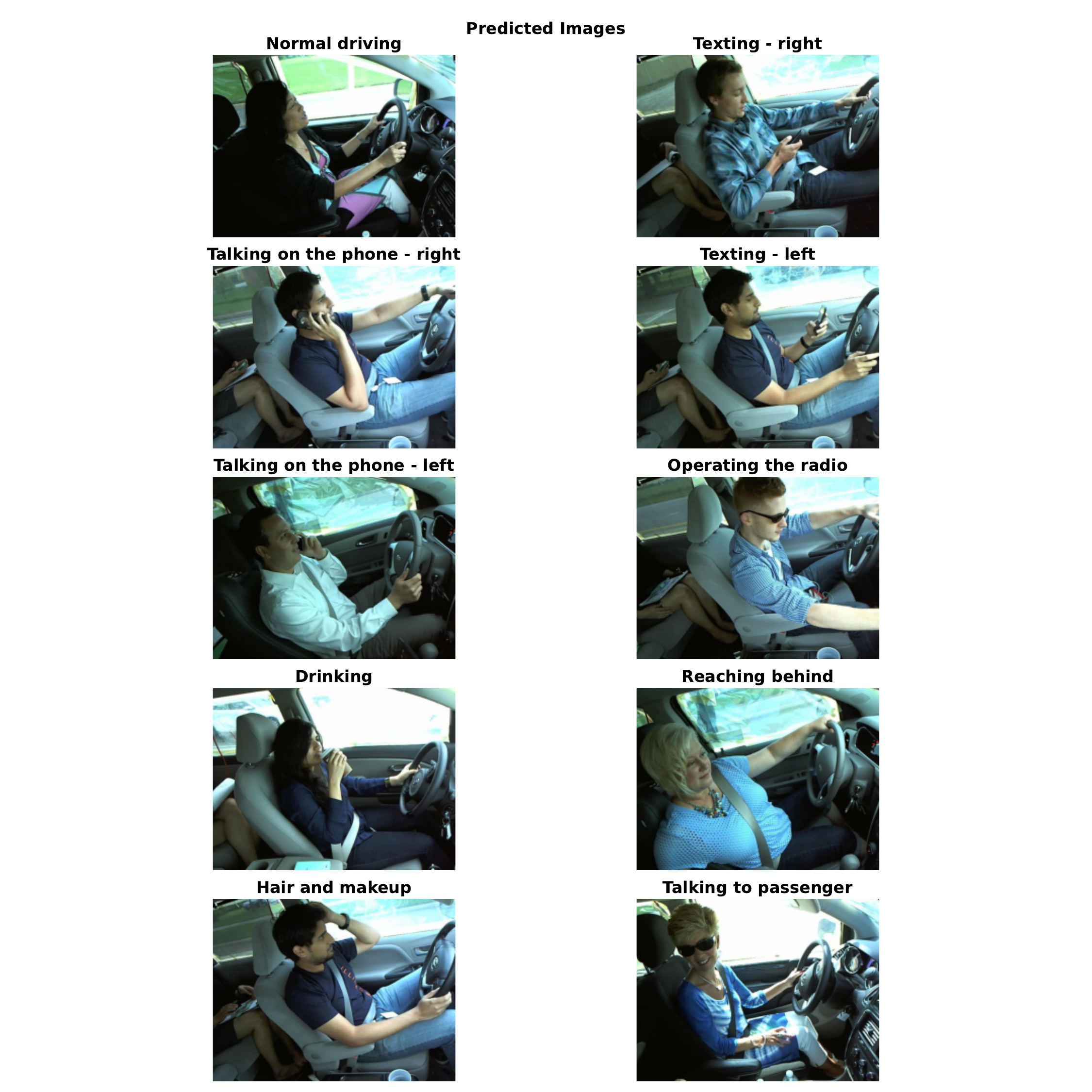}
\vspace{-0.7cm}
\caption{Predicted images illustrating the model's accuracy in identifying driving behaviors.}
\label{fig:Predicted_Images}
%\vspace{0cm} % Adjust this value as needed to
\end{figure}

To further illustrate the effectiveness of our model, we include a set of predicted images showcasing various driving behaviors, such as ``Normal driving," ``Texting - right" and ``Talking on the phone - right," with predictions that are consistent with the actual behaviors, demonstrating the applicability of the model in real-world scenarios as shown in Figure~\ref{fig:Predicted_Images}.

% \vspace{-0.4cm} % Adjust this value as needed to

\begin{figure}
\centering
\vspace{-0.4cm}
% L R B T
\includegraphics[scale=0.3, trim=80 0 0 22, clip]{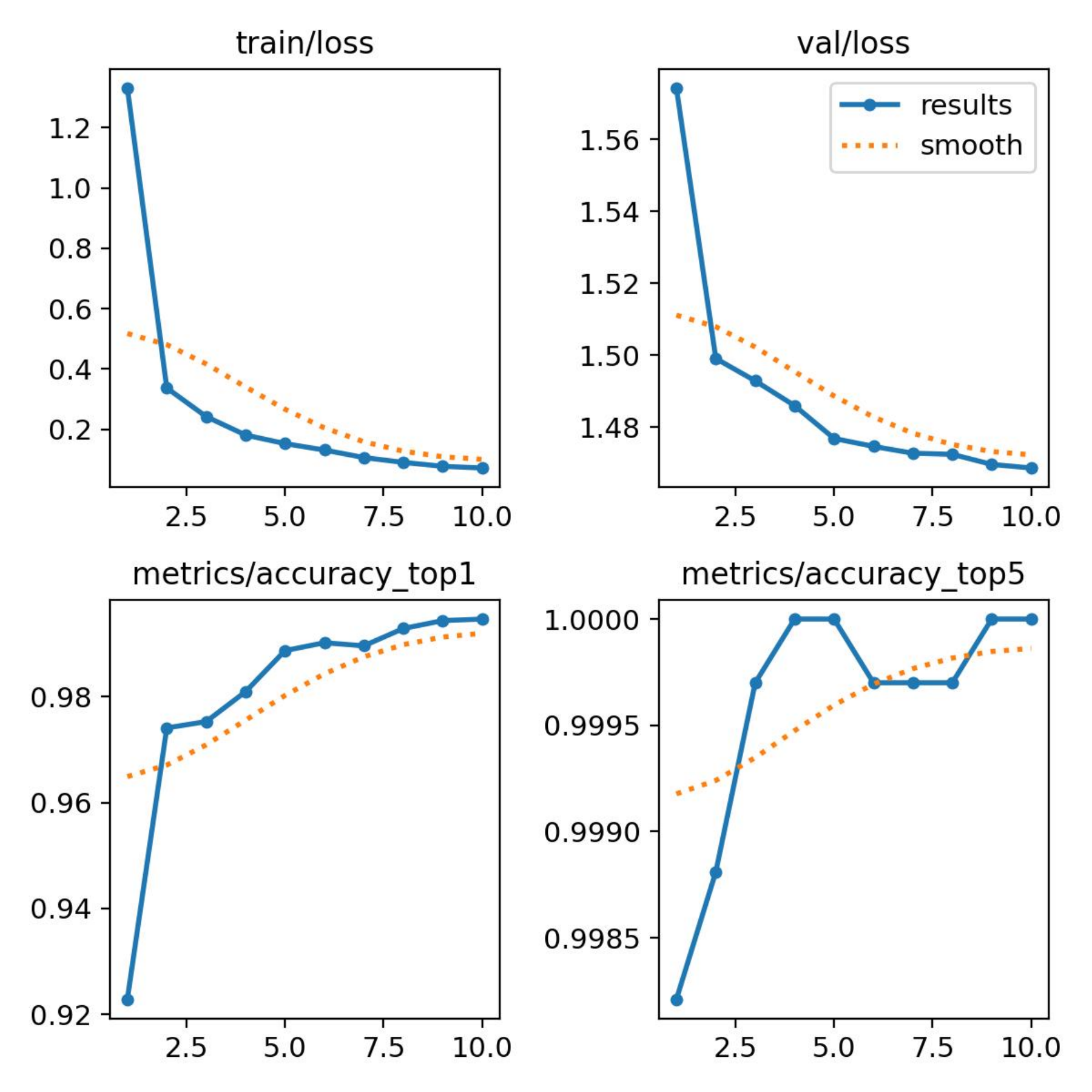}
% Adjust the trim values as needed
\vspace{-0.5cm}
\caption{Training and validation loss curves, along with Top-1 and Top-5 accuracy metrics over 10 epochs.}
\label{fig:Graghs}
\vspace{-0.3cm}
\end{figure}

Table~\ref{tab:class_numbers1} compares various models in the literature review with respect to the accuracy and the number of model parameters. The models cited include those by Hossain~\textit{et al.}~\cite{HOSSAIN2022200075}, Li~\textit{et al.}~\cite{10295832}, Abbas~\textit{et al.}~\cite{9318087}, Li~\textit{et al.}~\cite{9913644}, Subbulakshmi~\textit{et al.}~\cite{10353386}, and Masood~\textit{et al.}~\cite{Masood2020}. These models demonstrate a range of accuracies, from $91.04\%$ to $100\%$, and model sizes, from $1.36$ million to $491$ million parameters. 
The proposed algorithm, highlighted in the table, achieves an impressive accuracy of $99.46\%$ with a significantly smaller model size of $1.45$ million parameters. This represents an optimal trade-off between accuracy and model complexity. Although Li~\textit{ et al.}~\cite{9913644} reports a perfect accuracy of $100\%$, it comes at the cost of a substantially larger model size of $491$ million parameters, making it less practical for real-world applications with limited computational resources. In contrast, the proposed algorithm provides nearly equivalent accuracy with a fraction of the parameters, which improves its feasibility and efficiency. This analysis underscores the superiority of the proposed algorithm, which offers a balanced solution by maintaining high performance while being computationally efficient. Thus, it stands out as an optimal choice for applications where both accuracy and model size are critical considerations.

\begin{table}[tb]
\centering
\caption{Model Comparison in terms of accuracy and model size}
\label{tab:class_numbers1}
\begin{tabularx}{\columnwidth}{|c|X|c|}
\hline
Model & Accuracy & Model Parameters (millions) \\
\hline
Hossain~\textit{et al.} \cite{HOSSAIN2022200075} & 99.98\% & 3.5 \\
\hline
Li~\textit{et al.} \cite{10295832} & 91.04\% & 1.36 \\
\hline
Abbas~\textit{et al.} \cite{9318087} & 98\% & 317.4 \\
\hline
Li~\textit{et al.} \cite{9913644} & 100\% & 491 \\
\hline
Subbulakshmi~\textit{et al.} \cite{10353386} & 97.5\% & 5.3 \\
\hline
 Masood~\textit{et al.} \cite{Masood2020}& $99.57\%$ & $138$ \\
\hline
\textbf{Proposed algorithm} & \textbf{$99.46\%$} & \textbf{$1.45$} \\
\hline
\end{tabularx}
\end{table}

Our results highlight the proposed classification model accuracy and reliable performance in identifying different driving behaviors. The evaluation, supported by detailed metrics, confusion matrix analysis, and visual examples, confirms the robustness of the model and its potential for use in driver monitoring systems. We also present additional graphs that provide further insights into the model performance metrics and comparisons as depicted in Figure~\ref{fig:Graghs}. These curves represent the training and validation performance of a YOLOv8 classification model across epochs. The training loss curve shows that the training loss decreases significantly as the number of epochs increases, indicating that the model is learning and fitting the training data better over time. Similarly, the validation loss curve shows a decrease in validation loss over epochs, signifying that the model performance on unseen validation data is improving. This decrease in both training and validation loss is a positive sign that the model is generalizing well to new data. The Top-1 accuracy curve represents the proportion of times that the model top prediction (the one with the highest probability) is correct. This plot shows that Top-1 accuracy improves and stabilizes as training progresses, indicating that the model is becoming more accurate in its primary predictions. The top-5 accuracy curve represents the proportion of times the correct label is within the Top-5 predictions of the model. This plot shows a high Top-5 accuracy, approaching 100\%, which means that the model almost always includes the correct label among its top five predictions. The solid blue lines represent the actual values observed during training and validation, while the dotted orange lines represent smoothed versions of these metrics, helping to visualize the overall trends by reducing the noise in the data.

\section{Conclusion}

This study presents a real-time distracted driving detection system utilizing the Pretrained-YOLOv8 (P-YOLOv8) model, addressing both speed and accuracy challenges typically faced by traditional models. The results demonstrate the exceptional performance of P-YOLOv8 in image classification tasks, as evidenced by its application to the State Farm Driver Distraction Detection dataset. The model achieved an impressive accuracy of \(99.46\%\) while maintaining a compact parameter size of \(2.84 \, \text{MB}\) and \(1,451,098\) parameters. This demonstrates the model's computational efficiency and suitability for deployment on resource-constrained devices, such as those used in Tiny Machine Learning (TinyML). The P-YOLOv8 model, originally designed for object detection, has been adapted for efficient real-time processing in image classification. The P-YOLOv8 proves to be a viable alternative to traditional deep learning models, offering significant speed and computational cost advantages. The detailed analysis of the architecture, training, and performance benchmarks of P-YOLOv8 underscores its potential for practical applications to improve road safety through the detection of distracted driving behaviors. This approach, leveraging advances in the YOLO series, provides a balanced solution that maintains high accuracy while significantly reducing computational demands.

\section*{Acknowledgment}

The authors express their deepest gratitude to Prof.\ Francesco Fabiano for his input. Authors Mohamed R.\ Elshamy and Abdel-Hameed A.\ Badawy were partially supported by NSF grant 2219680.
% The preferred spelling of the word ``acknowledgment'' in America is without 
% an ``e'' after the ``g''. Avoid the stilted expression ``one of us (R. B. 
% G.) thanks $\ldots$''. Instead, try ``R. B. G. thanks$\ldots$''. Put sponsor 
% acknowledgments in the unnumbered footnote on the first page.

\bibliographystyle{plain}
\bibliography{References} % Use the single .bib file

\end{document}